\def\year{2019}\relax
\documentclass[letterpaper]{article} 
\usepackage{aaai19}  
\usepackage{times}  
\usepackage{helvet}  
\usepackage{courier}  
\usepackage{url}  
\usepackage{graphicx}  
\usepackage{subfig}

\usepackage{xspace}

\usepackage[dvipsnames]{xcolor}

\usepackage{xparse}

\newcommand{\etc}{\textit{etc.}\xspace}

\usepackage{multicol}
\usepackage{framed}

\renewcommand{\(}{\begin{columns}}
\renewcommand{\)}{\end{columns}}
\newcommand{\<}[1]{\begin{column}{#1}}
\renewcommand{\>}{\end{column}}

\usepackage{algorithm}
\usepackage{algorithmicx}
\usepackage[noend]{algpseudocode}
\usepackage{amsmath,amssymb,bm}
\usepackage{mathtools}

\usepackage{pifont}
\DeclareMathSymbol{\naf}{\mathord}{symbols}{"18}

\DeclarePairedDelimiterX{\infdivx}[2]{(}{)}{%
  #1\;\delimsize|\delimsize|\;#2%
}

\DeclareDocumentCommand \expectation { o m } {%
  \ensuremath{\mathbb{E}%
  \IfValueTF {#1} {%
    _{#1} \left[ #2 \right]%
  }{%
    \left[ #2 \right]%
  }%
  }\xspace%
}

\makeatletter
\newcommand*{\indep}{%
  \mathbin{%
    \mathpalette{\@indep}{}%
  }%
}
\newcommand*{\nindep}{%
  \mathbin{
    \mathpalette{\@indep}{/}%
  }%
}
\newcommand*{\@indep}[2]{%
  \sbox0{$#1\perp\m@th$}
  \sbox2{$#1=$}
  \sbox4{$#1\vcenter{}$}
  \rlap{\copy0}
  \dimen@=\dimexpr\ht2-\ht4-.2pt\relax
  \kern\dimen@
  \ifx\\#2\\%
  \else
    \hbox to \wd2{\hss$#1#2\m@th$\hss}%
    \kern-\wd2 %
  \fi
  \kern\dimen@
  \copy0 
}
\makeatother

\def\|#1{\ensuremath{\mathtt{#1}}}
\def\!#1{\ensuremath{\mathbf{#1}}}
\def\*#1{\ensuremath{\mathcal{#1}}}

\newcommand{\pvalue}{\ensuremath{p\text{-value}}\xspace}

\newcommand{\epmd}{\textsc{EP-md}\xspace}
\newcommand{\ep}{\textsc{Ep}\xspace}

\newcommand{\mort}{\textsc{mort}\xspace}
\newcommand{\los}{\textsc{LoS}\xspace}
\newcommand{\dd}{\textsc{DD}\xspace}

\newcommand{\mimic}{\textsc{MIMIC-III}\xspace}

\newcommand{\ehrs}{\textsc{EHR}s\xspace}
\newcommand{\icd}{\textsc{ICD}\xspace}

\newcommand{\sklearn}{\textsc{scikit-learn}\xspace}
\newcommand{\adam}{\textsc{Adam}\xspace}
\newcommand{\keras}{\textsc{Keras}\xspace}
\newcommand{\tensorflow}{\textsc{TensorFlow}\xspace}

\newcommand{\auroc}{\textsc{AuROC}\xspace}

\newcommand{\mae}{\textsc{MAE}\xspace}
\newcommand{\mcauroc}{\textsc{mc-AuROC}\xspace}

\newcommand{\combined}{\textsc{combined}\xspace}
\newcommand{\raw}{\textsc{raw}\xspace}
\newcommand{\embedded}{\textsc{embedded}\xspace}

\newcommand{\apptimeseries}{Appendix~A\xspace}
\newcommand{\appnotecategories}{Appendix~B\xspace}
\newcommand{\appdischargedestinations}{Appendix~C\xspace}
\newcommand{\apphyperparameterselection}{Appendix~D\xspace}
\newcommand{\appresulttables}{Appendix~E\xspace}
\newcommand{\appfeaturemaeloss}{Appendix~F\xspace}
\newcommand{\appfeatureimportance}{Appendix~G\xspace}

\begin{document}
%
\title{Learning Representations of Missing Data for\\Predicting Patient Outcomes}
\author{Brandon Malone, Alberto Garc{\'i}a-Dur{\'a}n, and Mathias Niepert}
\maketitle

\begin{abstract}
    Extracting actionable insight from Electronic Health Records (\ehrs) poses several challenges for traditional
    machine learning approaches. Patients are often missing data relative to each
    other; the data comes in a variety of modalities, such as multivariate time
    series, free text, and categorical demographic information; important
    relationships among patients can be difficult to detect; and many others. In
    this work, we propose a novel approach to address these first three challenges
    using a representation learning scheme based on message passing. We show
    that our proposed approach is competitive with or outperforms the state of the art
     for predicting in-hospital mortality (binary classification), the length of hospital visits 
     (regression) and the discharge destination (multiclass classification). 
\end{abstract}

\section{Introduction}
\label{sec:introduction}

Healthcare is an integral service in modern societies. Improving its quality and efficiency has proven to be difficult and costly. Electronic health records (\ehrs)
provide a wealth of information carrying the potential to improve treatment quality and patient outcomes. Extracting
useful and actionable medical insights from \ehrs, however, poses several challenges both to traditional statistical
and machine learning techniques.

First, not all patients receive the same set of laboratory tests,
examinations, consultations, \etc, while they are at the hospital. Thus,
many patients have \emph{missing and incomplete data}, relative to other patients.
Second,
the various medical conditions and the corresponding treatment activities yield different kinds of data. For example,
a blood oxygen saturation sensor may collect numeric values for a given amount
of time at fixed frequency, while a consultation with a physician may produce
only free text notes of the physician's interpretation. That is, there are
multiple \emph{modalities} (or \emph{attribute types}) of data, and these modalities have variations due to 
a number of external factors.
Third, patients may share important relationships which are not easily
captured in typical data representations. For example, family members often
share a similar genetic background. Likewise, patients with similar initial
diagnoses may share underlying characteristics which are difficult to
capture in traditional models. Thus, it is important to explicitly model
\emph{relationships} among patients that capture some form of disease or treatment affinity.

Recent work on representation learning for graph-structured data 
has specifically addressed the last two problems albeit in different domains. In that line of work, edges in an \emph{affinity graph} connect
 instances deemed similar in some way, and a message passing scheme is used
to learn representations of the data modalities associated with each instance.
While several message-passing approaches to graph representation learning have been proposed~\cite{Kipf2017,hamilton2017inductive,GilmerSRVD17}, we base our work on embedding propagation (\ep), a method that is both unsupervised and learns a representation specific to each data modality~\cite{GarciaDuran2017}.

\ep has several characteristics that make it well-suited for the problem of learning from medical records. First, it provides a  method for learning and combining representations, or \emph{embeddings}, of the various data modalities typically found in EHRs. For instance, it is possible to simultaneously learn embeddings of free-text notes  and to combine these with embeddings learned for sensor time series data.
Second, as we will propose in this work, due to its unsupervised reconstruction loss, \ep allows us to learn a vector representation for every data modality and every patient, even if that particular data modality is not observed at all for some of the patients.
Third, \ep learns representations for each data modality independently; thus, it is possible to distinguish between the influence of these independently learned modality representations on the predictions.

Intuitively, embedding propagation learns data representations such that representations of nodes close in the graph, that is, nodes similar in one or several ways, are more similar to each other than those of nodes far away in the graph. The representations can then be used in traditional downstream machine learning tasks. In the context of \ehrs, patients are modeled as nodes in the graph\footnote{Specifically, patient episodes in an intensive care unit (ICU) are modeled as nodes; for ease of exposition, we use ``patient'', ``episode'' and ``node'' interchangeably in this work. The meaning is clear from the context.}, and similarity relationships between patients  are modeled with edges. 

In this work, we extend the embedding propagation framework to account for missing data. In particular,
for each data modality, we learn \emph{two} representations for each
patient; the first representation carries information about the observed data, while the
second representation is learned to account for missing data. Learning an explicit representation of missing data within the graph-based learning framework has the advantage of propagating representations of missing data throughout the graph such that these representations are also informed by representations of neighboring patients. Combining these learned feature representations gives
a complete representation for downstream tasks.

We use a recently-introduced benchmark based on the \mimic
dataset~\cite{Johnson2016a} to evaluate the proposed approach and show that it is competitive with the state of the art
when using a single data modality, numeric time series observations. After
augmenting the data with additional data modalities, including free text from
physicians' notes and demographic information, we outperform the state of the
art on two of the three tasks we consider in this work, namely, length of hospital stay
and discharge destination prediction.

\begin{figure}%
    \centering
    \subfloat[]{{\includegraphics[width=0.22\textwidth]{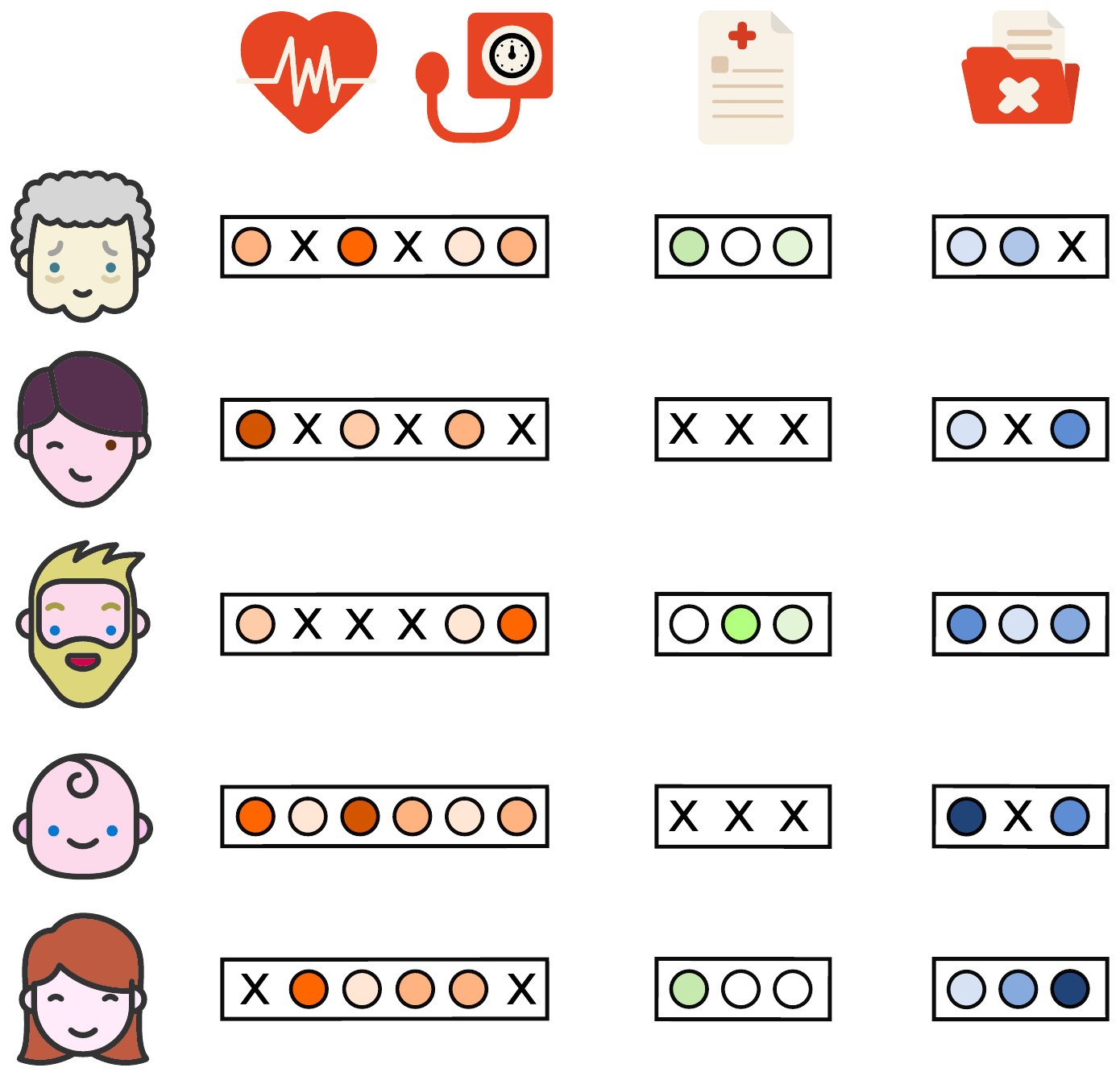} }}%
    \hspace{3mm}
    \subfloat[]{{\includegraphics[width=0.22\textwidth]{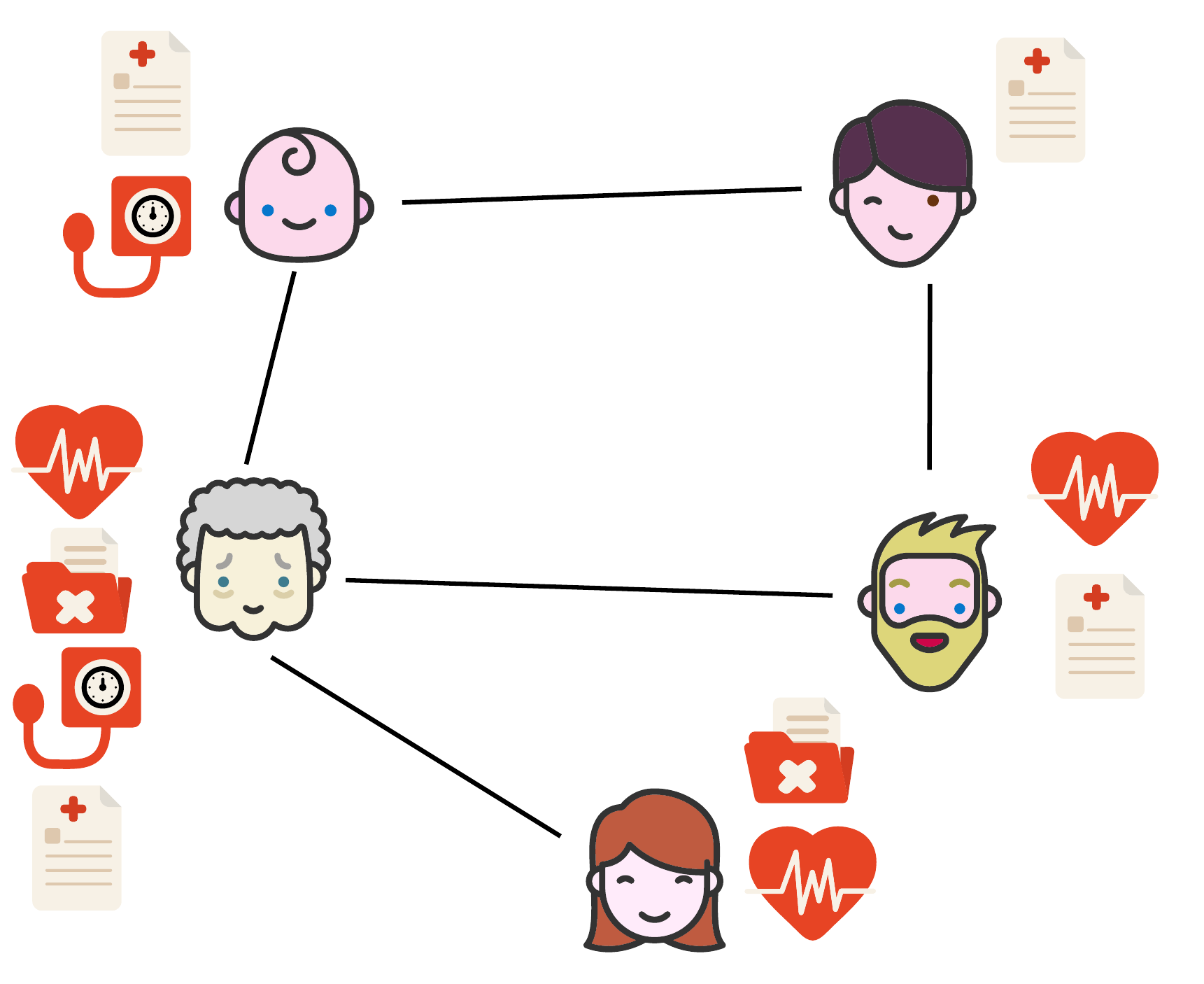} }}%
    \caption{\label{fig:example} (a) Illustration of patients and the corresponding time-series, text, and demographic data. Missing data is indicated with an $\mathtt{x}$. (b) A patient affinity graph constructed with a subset of the patient attributes. The graph structure is leveraged in the EP framework. }%
\end{figure}

The rest of this paper is structured as follows. In Section~\ref{sec:data}, we
describe the data that we use in this work. Section~\ref{sec:ep} introduces \ep
and describes our novel contribution for handling missingness within it. We describe
our experimental design and empirical results in Section~\ref{sec:experiments}.
Section~\ref{sec:related-work} places our contribution in context with related
work, while Section~\ref{sec:discussion} concludes the paper.

\section{Data}
\label{sec:data}

In this work, we use the \mimic~\cite{Johnson2016a} (Medical Information Mart for Intensive Care III) benchmark datasets
created by \cite{Harutyunyan2017} which contain information about patient intensive care unit (ICU) visits, or \emph{episodes}. However, these benchmarks include only
time series data. In order to demonstrate the efficacy of our proposed approach
in a multimodal context, we augment the benchmarks to include textual data and
(categorical) demographic data. This section describes our dataset 
as well as the specific tasks we consider.

\subsection{Adding Additional Modalities}
\label{sec:adding-modalities}

The original benchmark dataset 
includes $17$ time series variables (see \apptimeseries in the Supplementary Material), sampled at
varying frequencies and many of which are missing for some patients (please see
\cite{Harutyunyan2017} for more details). Further, each admission includes a
\|{SUBJECT\_ID} and \|{EPISODE} which link it back to the complete \mimic
dataset. We use these foreign keys to augment the time series variables with
text (from the \|{NOTEEVENTS} table) and demographic (from the \|{ADMISSIONS}
and \|{PATIENTS} tables) data. In both cases, we ensure to only use data
available at the time of prediction.\footnote{We will make the scripts used to
create the dataset publicly-available.} For example, we never use "Discharge
Summaries" to predict mortality.

\subsection{Preprocessing}
\label{sec:preprocessing}

We extracted features from the time series data as previously
described~\cite{Harutyunyan2017}. Briefly, we create seven sets of observations
from each time series (the entire sequence, and the first and last $10\%$, $25\%$
and $50\%$ segments of the sequence). We then calculate nine features for each
segment (count, minimum, maximum, mean, standard deviation, skew, kurtosis,
median, and max absolute deviation). Thus, in total, for the time series data we
have $7 \times 9 = 63$ features.\footnote{This is slightly higher
than the prior work because we also include median and max absolute deviation
features since they are robust against outliers.} Finally, we standardize all
observations such that each feature has a mean of $0$ and a variance of $1$
for the observed values across all patients in the training set.

The notes were first partitioned based on their \|{CATEGORY} (see \appnotecategories) into $6$ note types:
nursing, radiology, respitory, ecg, echo and other. As mentioned above, we
never use the discharge summaries in this work. We then convert each note into
a bag of words representation; we discard all words which appear in less than
$0.1\%$ or more than $90\%$ of the notes. The concatenated notes for each
note type are used as the text observations for each patient. That is, each
patient has $6$ bag-of-word text features, one for each type.

Further, we extract admission and demographic information about each episode
from the respective tables. In particular, we collect the \|{ADMISSION\_TYPE}
(``urgent'', ``elective'', \etc), \|{ADMISSION\_LOCATION} (``emergency room'',
\etc), \|{DIAGNOSIS} (``a preliminary, free text diagnosis \dots [which is] usually
assigned by the admitting clinician and does not use a systematic ontology''
\footnote{{https://mimic.physionet.org/mimictables/admissions/}} about the
admission. Each of these three fields contains text. We also collect the
patient's ethnicity, gender, age, insurance,
and marital status.

Figure~\ref{fig:example} gives an example of the dataset after preprocessing
and graph construction (see Section~\ref{sec:experiments} for details about
graph construction).

\subsection{Patient Outcome Labels}
\label{sec:patient-outcome-labels}

In this work, we consider three prediction tasks: in-hospital
mortality (\mort), length of stay (\los), and discharge destination (\dd). In
all cases, we make predictions $48$ hours after admission to the ICU. We adopt
the same semantics for the starting time of an episode as \cite{Harutyunyan2017}.

For \mort, this is exactly the same problem in the original benchmarks. For \los,
prediction at $48$ has previously been used~\cite{Silva2012}.

Several medical studies have considered the \dd task~\cite{Mauthe1996,Brauer2008,Wee1999}; 
however, these approaches have relied on traditional clinical scores such as the
FIM or Berg balance score.
To the best of our knowledge, the \dd task has not been extensively considered
in the machine learning literature before. 
In contrast to the binary \mort classification task, this is a multiclass
classification problem. In particular, the set of patients in the \mimic
in this study have $6$ discharge destinations
(after grouping; see \appdischargedestinations). Conceptually, we
consider \dd as a more fine-grained proxy for the eventual patient outcome
compared to \mort.

\section{EP for Learning Patient Representations}
\label{sec:ep}

The general learning framework of \ep~\cite{GarciaDuran2017} proceeds in two separate steps. We use the term \emph{attribute} to refer to a particular node feature. Moreover, an \emph{attribute type} (or \emph{data modality}) is either one node attribute or a set of node attributes grouped together.

In a first step, \ep learns a vector representation for every attribute type by passing messages along the edges of the affinity graph. In the context of learning from EHRs, for instance, one attribute type consists of time series data recorded for ICU patients. The attribute types used in this work are: (1) a group of time-series data from ICU measurements; (2) words of free-text notes; and (3) a group of categorical features of demographic data. As described in Section~\ref{sec:preprocessing}, we have several attributes of each type.

In a second step, \ep learns patient representations by combining the previous learned attribute representations. For instance, the method combines the representations learned for the physicians' notes with the representations learned for the time series data. These combined representations are then used in the downstream prediction tasks.

\subsection{Attribute Type Representation Learning}
\label{sec:label-learning}

Each attribute type $i$ is associated with a domain $\mathcal{D}_i$ of possible values. For instance, an attribute type consisting of $N$ numerical attributes has the domain $\mathbb{R}^{N}$, while an attribute type modeling text data has the domain $\mathbb{R}^{|T|}$ where $T$ is the vocabulary. 
For each attribute type we choose a suitable encoding function $\mathtt{f}$. This encoding function $\mathtt{f}$ is parameterized and maps every $\mathbf{x} \in \mathcal{D}_i$ to its vector representation $\mathbf{x}' \in \mathbb{R}^{d_i}$, that is, $\mathbf{x}' = \mathtt{f}_i(\mathbf{x})$. These encoding functions have to be differentiable to update their parameters during learning. Concretely, in that work, a single dense layer is used as the encoding function for each label type. For the time series-derived numeric attribute, this amounts to a standard matrix multiplication, where as for the text and categorical attribute types, this amounts to an embedding lookup table.

The functions $\mathtt{l}_i: V \rightarrow 2^{\mathcal{D}_i}$ map every vertex $v$ in the graph to a (possibly empty) set of vectors $\mathbf{x} \in \mathcal{D}_i$. We write $\mathtt{l}(v) = \bigcup_{i}\mathtt{l}_i(v)$ for the set of all vectors of attribute type $i$ associated with vertex $v$. Moreover, we write $\mathtt{l}_i(\mathbf{N}(v)) = \lbrace{\mathtt{l}_i(u)} \mid u \in \mathbf{N}(v)\rbrace$ for the multiset of vectors of attribute type $i$ associated with the neighbors of vertex $v$.
\ep learns a vector representation for each attribute type of the problem. 

\begin{itemize}
\item We write $\mathbf{h}_{i}(v)$ to denote the current vector representation of attribute type $i$ for node $v$. It is computed as follows:
\begin{equation}
\small
\mathbf{h}_{i}(v) = \mathtt{g}_i\left(\lbrace \mathtt{f}_i\left(\mathbf{x}\right) \mid \mathbf{x} \in \mathtt{l}_i(v)\rbrace\right).
\end{equation}
\item We write $\widetilde{\mathbf{h}}_{i}(v)$ to denote the reconstruction of the representation of attribute type $i$ for node $v$. $\widetilde{\mathbf{h}}_{i}(v)$ is computed from the attribute type representations of \textit{v}'s neighbors in the graph. It is computed as follows:
\begin{equation}
\small
\widetilde{\mathbf{h}}_{i}(v) = \widetilde{\mathtt{g}}_i\left(\lbrace\mathtt{f}_i\left(\mathbf{x}\right) \mid \mathbf{x} \in \mathtt{l}_i(\mathbf{N}(v))\rbrace\right),
\end{equation}
\end{itemize}
where $\mathtt{g}_i$ and $\widetilde{\mathtt{g}}_i$ are aggregation functions that map a multiset of of $d_i$-dimensional embeddings to a single $d_i$-dimensional embedding. These aggregation functions can be parameterized but are often parameter-free aggregation functions such as the elementwise average or maximum. 

The core idea of \textsc{Ep} is to make the attribute type representation and its reconstruction similar for each attribute type and each node in the graph. In other words, \ep learns attribute type representations such that the distance between $\mathbf{h}_{i}(v)$ and $\widetilde{\mathbf{h}}_{i}(v)$ is small. More formally, for all attribute types \ep minimizes the following loss
\begin{equation}
\small
\label{Ep-margin-ranking-loss}
\mathcal{L}_i =  \sum_{v \in V}  \sum_{u \in V\setminus \{v\}} 
\hspace{-2mm} \left[ \gamma +  \mathtt{d}_i\left(\widetilde{\mathbf{h}}_{i}(v), \mathbf{h}_{i}(v)\right) - \mathtt{d}_i\left(\widetilde{\mathbf{h}}_{i}(v), \mathbf{h}_{i}(u)\right)\right]_{+}
\end{equation}
where $\mathtt{d}_i$ is the Euclidean distance, $[x]_{+}$ is the positive part of $x$, and $\gamma > 0$ is a margin hyperparameter.

In practice, it is not feasible to evaluate the inner summation;
thus, it is approximated by sampling a single random node $u$ which is different than $v$.

The margin-based loss defined in Equation \ref{Ep-margin-ranking-loss} updates the parameters (i.e., embedding functions and functions $\mathtt{g}_i$ and $\widetilde{\mathtt{g}}_i$ in case they are chosen to be parametric) if the distance between $\mathbf{h}_{i}(v)$ and  $\widetilde{\mathbf{h}}_{i}(v)$ plus a margin is not smaller than the distance between $\mathbf{h}_{i}(u)$ and  $\widetilde{\mathbf{h}}_{i}(v)$. Intuitively, for each patient node $v$ and attribute type $i$, the vector representation reconstructed from the embeddings of patient nodes neighboring $v$ are learned to be more similar to the embedding of attribute type $i$ for $v$ than to the embedding of attribute type $i$ of a random patient node in the graph.

The generic working of the attribute type representation learning stage is as follows: in each propagation step, for each node $v$ in the graph, randomly sample a node $u$ uniformly from the set of all nodes in the graph; then, for each attribute type $i$, the embeddings $\widetilde{\mathbf{h}}_{i}(v)$, $\mathbf{h}_{i}(v)$ and $\mathbf{h}_{i}(u)$ are computed.
Finally, parameters of the model are updated based on the loss defined in Equation~(\ref{Ep-margin-ranking-loss}).

The functions $\mathtt{g}_i(\mathcal{H})$ and $\widetilde{\mathtt{g}}_i(\mathcal{H})$ are given as
\begin{small}
$$\mathtt{g}_i(\mathcal{H}) = \widetilde{\mathtt{g}}_i(\mathcal{H}) = \frac{1}{|\mathcal{H}|}\sum_{\mathbf{h} \in \mathcal{H}} \mathbf{h}$$

\end{small}
for attribute types $i$ and sets of embedding vectors $\mathcal{H}$. 

\begin{figure*}[t!]
\centering
\includegraphics[width=0.7\textwidth]{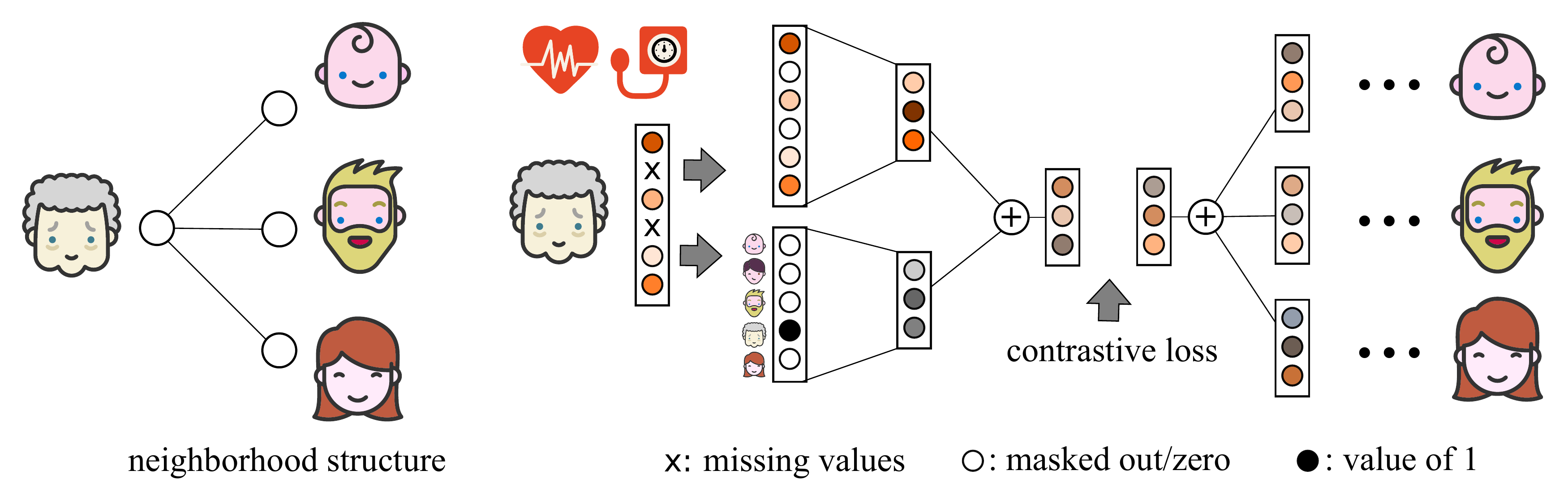}
\caption{\label{fig-ep-patient-missing} Illustration of the embedding propagation framework for missing data.}
\end{figure*}

\subsection{Learning Representations for Missing Data}
\label{sec:lr-missing-data}

Two characteristics of \ep make it suitable for the problem of learning representations for missing data. First, it supports an arbitrary number of attribute types, and one can learn missing data representations tailored to attribute types. Second, \ep's learning principle is based on reconstructing each node's attribute representation from neighboring nodes' attribute representations, and this makes it highly suitable for settings wherein a number of nodes have missing data.
During training, \ep learns how to reconstruct the missing data representation based on a contrastive loss between representations of existing attribute types and, therefore, can learn how to reconstruct a representation when data is missing. 

For every attribute type $i$ and for every node $v \in V = \{1, ..., |V|\}$ of the graph we have an input feature vector $\mathbf{x}$.
Based on this vector, we create two feature vectors.
The first feature vector $\mathbf{x}_{1}$ is identical to $\mathbf{x}$ except that all missing values are masked out.
The second feature vector $\mathbf{x}_{2} \in \mathbb{R}^{|V|}$ is either (1) all zeros if there are no missing attribute values for attribute type $i$ \emph{or} (2) all zeros except for the position $v$ which is set to $1$. The vector $\mathbf{x}_{2}$ indicates whether data is missing and and is used to learn a latent representation for nodes with missing data.  These two vectors are then fed into two encoding functions $\mathtt{f}^{1}_i$ and $\mathtt{f}^{2}_i$ whose parameters are learned independently from each other. For each input, the output vectors of the encoding functions are added element-wise, and the resulting vector is used in the margin-based contrastive loss function. Therefore, for every attribute type $i$ we have $\mathbf{x}' = \mathtt{f}_i(\mathbf{x}) = \mathtt{f}^{1}_i(\mathbf{x}_{1}) + \mathtt{f}^{2}_i(\mathbf{x}_{2})=  \mathbf{x}_{1}^{T} \mathbf{W}^{1}_i + \mathbf{x}_{2}^{T} \mathbf{W}^{2}_i$. The $\mathtt{f}^{2}_i$ encoding functions amount to embedding lookup tables. We refer to this instance of \ep as \epmd (for ``EP-missing data'').

Figure~\ref{fig-ep-patient-missing} illustrates our approach to learning separate representations for observed and missing attribute labels. Since the contrastive loss compares representations that are computed both based on missing and observed data, the two representations influence each other. If for some node $v$ and attribute type $i$  we have missing data, the representation of that missing data is influenced by the representations of observed and missing data of attribute type $i$ by neighboring nodes. Thus, the missing data representations are also propagated during the message passing.

As mentioned in Section~\ref{sec:preprocessing}, we standardize features to have a mean of $0$;
$\mathbf{x}_{2}$ allows \epmd to distinguish between observed $0$s and missing values which have been masked.

\paragraph{Missing data mechanisms}
Missing data is often assumed to result from mechanisms such as Missing (Completely) at Random (MAR and MCAR) or Not Missing at Random (NMAR)~\cite{Rubin1976}.
In contrast to methods like multiple imputation by chained equations~\cite{Azur2011}, \epmd does not estimate (representations of) missing attributes based on other observed attributes;
rather, it leverages the affinity graph to learn representations based on the same attribute observed for similar nodes.
Thus, \epmd does not assume an explicit missingness mechanism.
We leave exploration of the statistical missingness properties of \epmd to future work.

\subsection{Generating Patient Representations}
\label{sec:patient-learning}

Once the learning of attribute type representations has finished, \ep computes a vector representation for each patient node $v$ from the vector representations of $v$'s attribute types.
In this work, the patient representation is the concatenation of the attribute type representations:
\begin{equation}
\small
\mathbf{v} = \mathtt{concat}\left[\mathbf{h}_{1}(v), ..., \mathbf{h}_{k}(v) \right].
\end{equation}
Since we have modeled missing data explicitly, the latent representations $\mathbf{h}_{i}(v)$ exist for every node in the graph and every attribute type $i$. 
We also evaluate a patient representation built from \emph{both} the raw features and the learned representations:
\begin{equation}
\small
\mathbf{v} = \mathtt{concat}\left[\mathbf{h}_{1}(v), ..., \mathbf{h}_{k}(v), \mathbf{x}_{1}(v), ..., \mathbf{x}_{k}(v) \right],
\end{equation}
where $\mathbf{x}_{i}(v)$ is the raw input for attribute type $i$ and node $v$.


\section{Experiments}
\label{sec:experiments}

We first describe our experimental design. We then present
and discuss our results on the three patient outcome prediction tasks.

\paragraph{Hyperparameters and computing environment}

For \epmd, we use $200$ iterations of training with mini-batches of size $256$. We embed
all attributes in a $32$-dimensional space. We
set the margin $\gamma = 5$, and we solved the \epmd optimization problem using
\adam with a learning rate of $1e-3$. \epmd was implemented in \keras using
\tensorflow as the backend. All experiments were run on computers
with $128$GB RAM, four quad-core $2.8$ GHz CPU, and a TitanX GPU.
The computing cluster was in a shared environment, so we do not report
running times in this evaluation. In all cases, though, the running times were
modest, ranging from a few minutes up to about an hour for learning embeddings for
all episodes for a single attribute type.

\paragraph{Baseline and downstream models}

Prior work has shown that linear models perform
admirably compared to much more sophisticated methods on these and similar
benchmark problems~\cite{Lipton2016a,Harutyunyan2017,Barnes2016}. Thus, we use (multi-class) logistic regression as a
base model for \mort and \dd and ridge regression as the base model for \los.
Both implementations are from \sklearn~\cite{Pedregosa2011}.

We leave a more comprehensive comparison to complex models such as long short-term memory networks (LSTMs) as future work.
However, we do include comparison to previously-published performance of an
LSTM~\cite{Harutyunyan2017} on \mort.
For a baseline, missing values are replaced with the mean of the respective feature.
This completed data is then used to train the base model. 
We compare the baselines to learning embeddings with \epmd followed 
by the same supervised learning approach (either logistic or ridge regression).
We select all hyperparameters based on cross-validation (please see
\apphyperparameterselection for more details).
\ep has been shown to outperform methods like \textsc{DeepWalk}~\cite{perozzi2014deepwalk} and graph convolutional networks~\cite{Kipf2017};
thus, we do not compare to those methods in this work.


\paragraph{Construction of the affinity graph}
An integral component of \epmd is the affinity graph; it gives a mechanism
to capture domain knowledge about the similarity of the nodes to which
traditional machine learning approaches do not have access. In this work,
since the goal is always to make a prediction about the outcome of an episode,
each node in the graph corresponds to an episode.

As described in Section~\ref{sec:preprocessing}, we extract the admission type,
location, and initial diagnosis about each episode. We base the construction of
our graph on the text from these fields. In particular, we concatenate the
three fields to form a ``document'' for each episode. Next, we use
\|{fastText}\footnote{\|{https://github.com/facebookresearch/fastText}} to
learn custom word embeddings via skipgrams from this corpus. We then use these
to calculate a single embedding (a ``sentence vector''; please see the software
documentation for more details) for each episode.
We then define the similarity between two episodes $i$ and $j$ as
$s_{i,j} = \exp{\{-d_{i,j}\}}\text{,}$ where $d_{i,j}$ is the Euclidean distance between the respective 
embeddings. Finally, we connect all pairs of episodes for which $s_{i,j} > 0.9$.

\paragraph{Data partitioning}

\cite{Harutyunyan2017} created a standard training and testing set split
of episodes. The training set includes
$17~869$ episodes; the testing set includes $3~233$ episodes. While modest
in comparison to some datasets considered in the machine learning community, this dataset is huge compared to many datasets presented in
the medical literature. Thus, we also consider the ability to generalize to the
test set when outcome information is available for only smaller numbers of training
episodes.
In particular, we draw subsets of varying size (see Figure~\ref{fig:results}) and only observe the class label for those episodes. As
described in Section~\ref{sec:lr-missing-data}, \epmd can still take advantage
of the unlabeled episodes. On the other hand, completely supervised methods
can only use the episodes with class labels.
We note that our predictions are always made $48$ hours after admission. We ensure
that our learning only considers information that would be available in such a
realistic scenario.

\paragraph{Evaluation metrics}

We use standard metrics  for evaluation. We use
the area under the receiver operating characteristic curve (\auroc) to evaluate
\mort, mean absolute error (\mae) to evaluate \los, and a multi-class
generalization of \auroc (\mcauroc)~\cite{Hand2001} for \dd.

\subsection{Results}
\label{sec:results}

\begin{figure*}
\centering
    
    \center
    
    \vspace{-2mm}
        
        \textbf{Time series modality only}
    

    \includegraphics[width=0.3\textwidth,keepaspectratio]{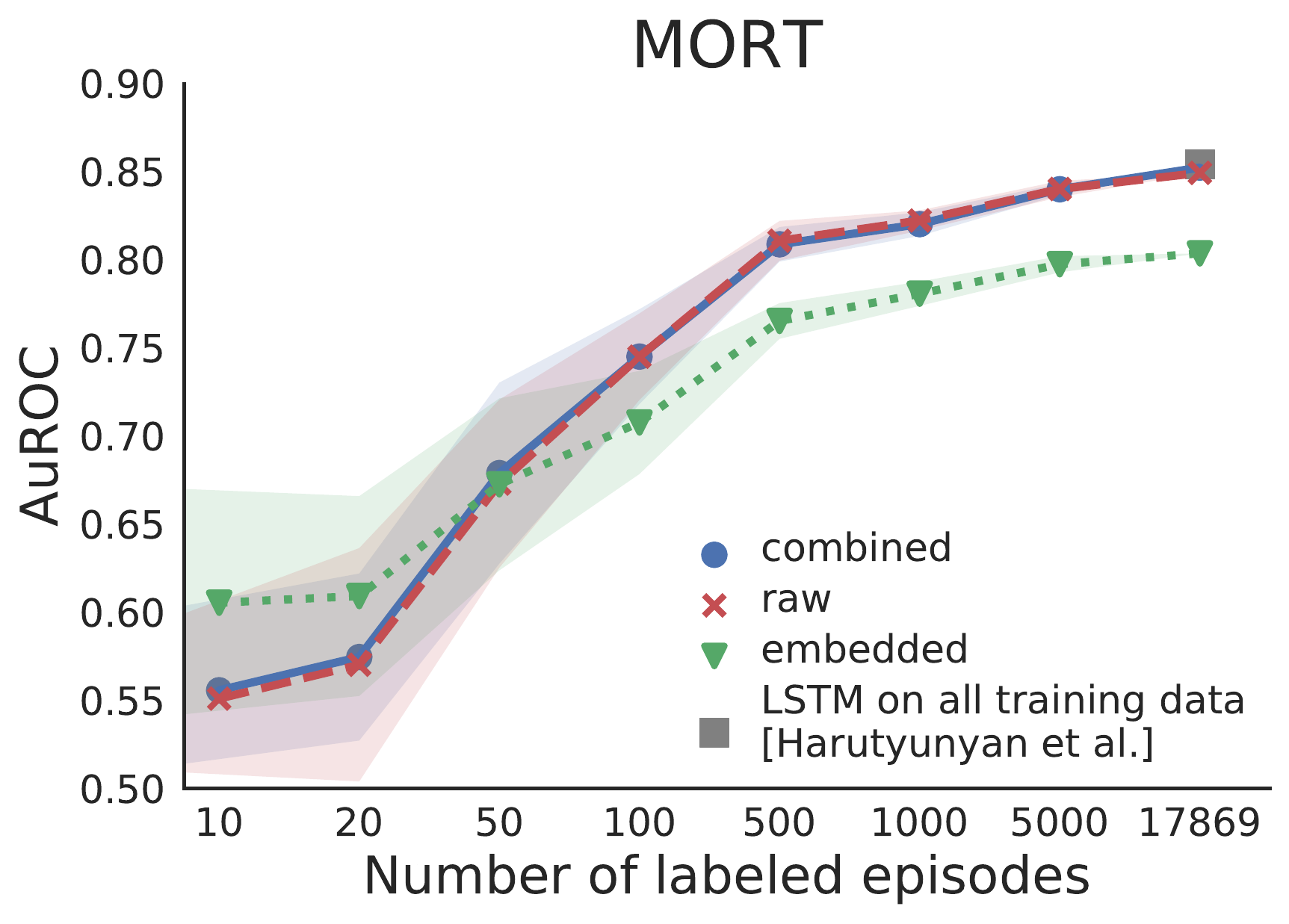}%
    \hspace{5mm}
    \includegraphics[width=0.3\textwidth,keepaspectratio]{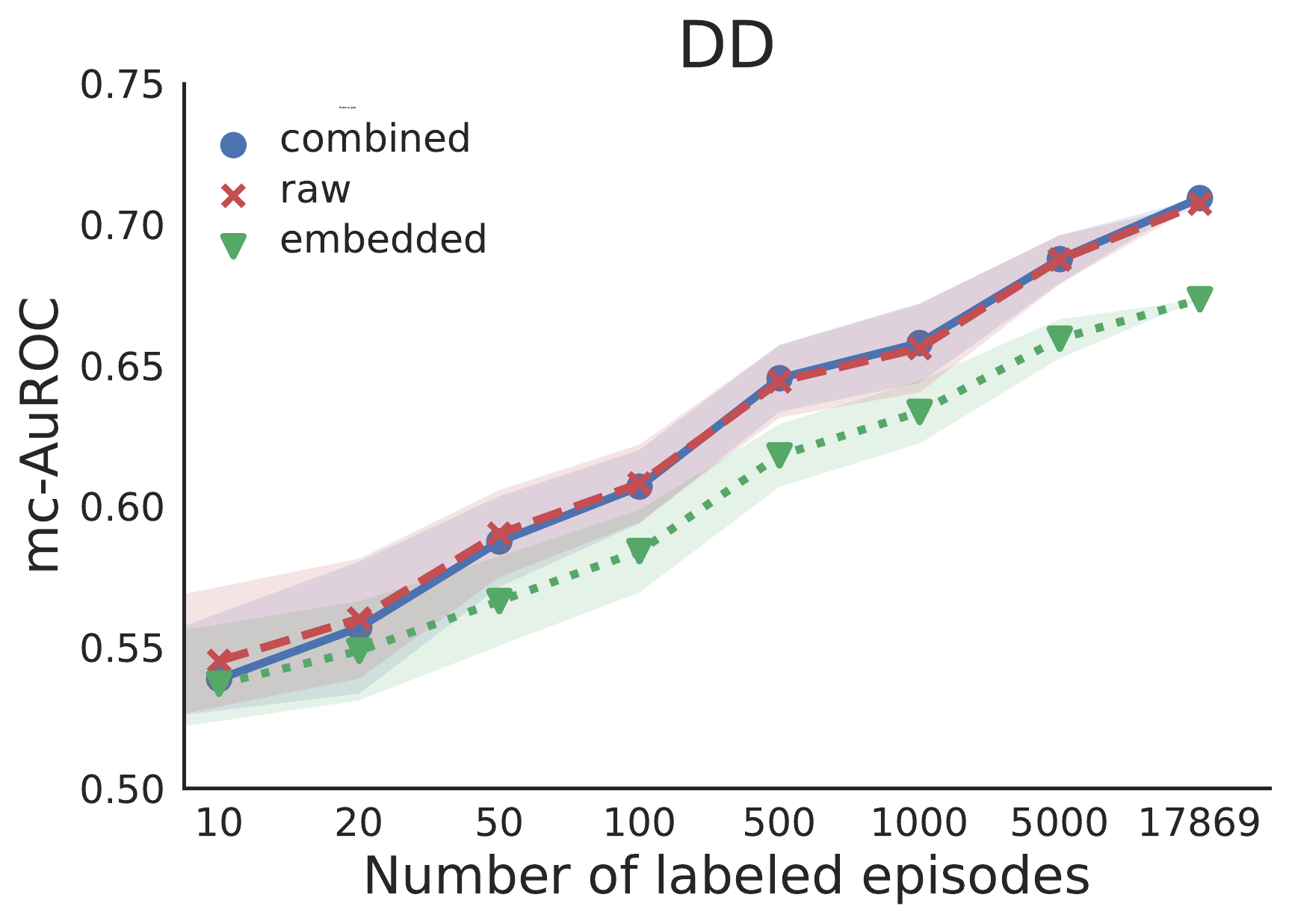}%
    \hspace{5mm}
    \includegraphics[width=0.3\textwidth,keepaspectratio]{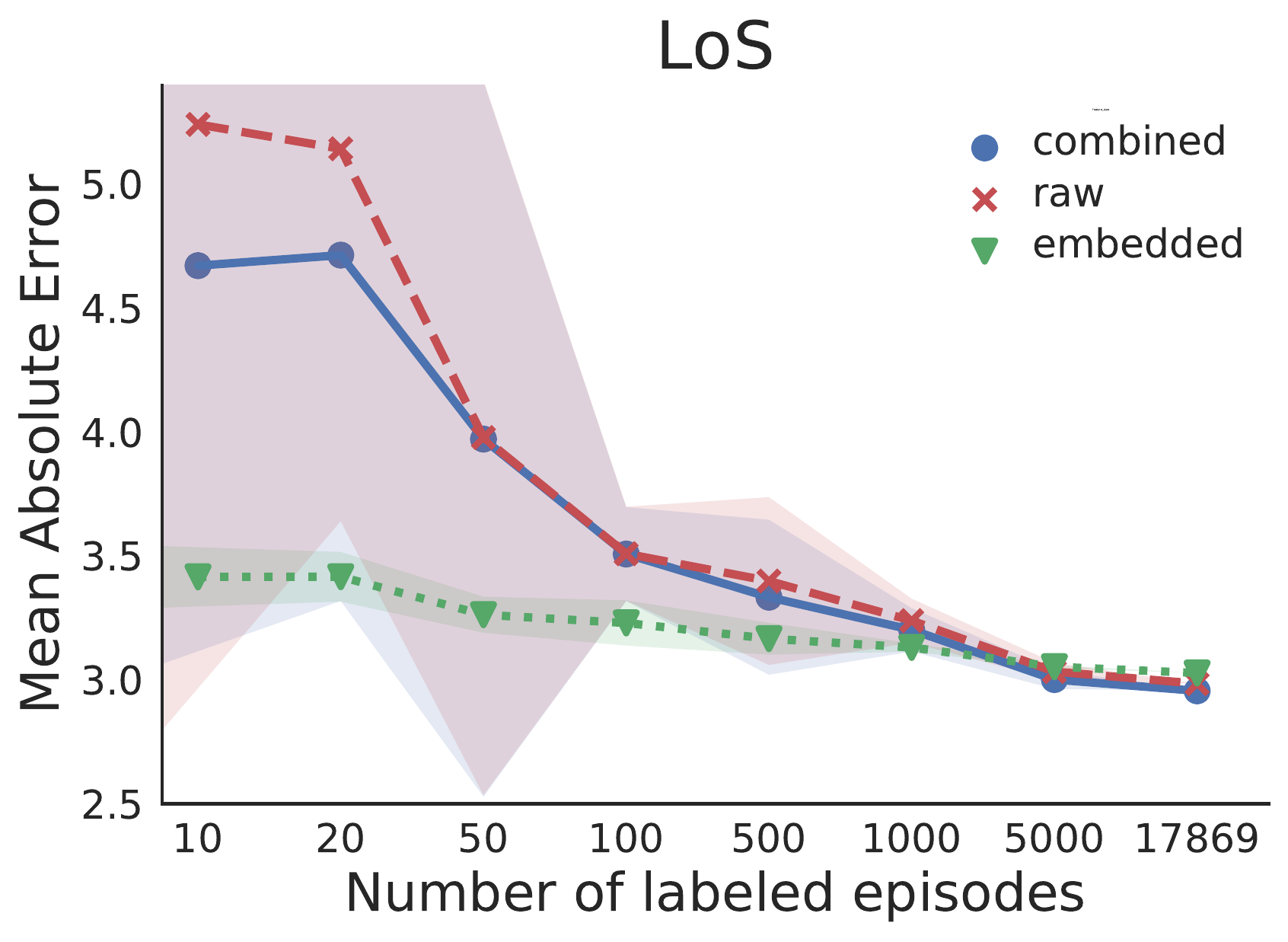}%
    
    
    \textbf{All data modalities}
    

    \includegraphics[width=0.3\textwidth,keepaspectratio]{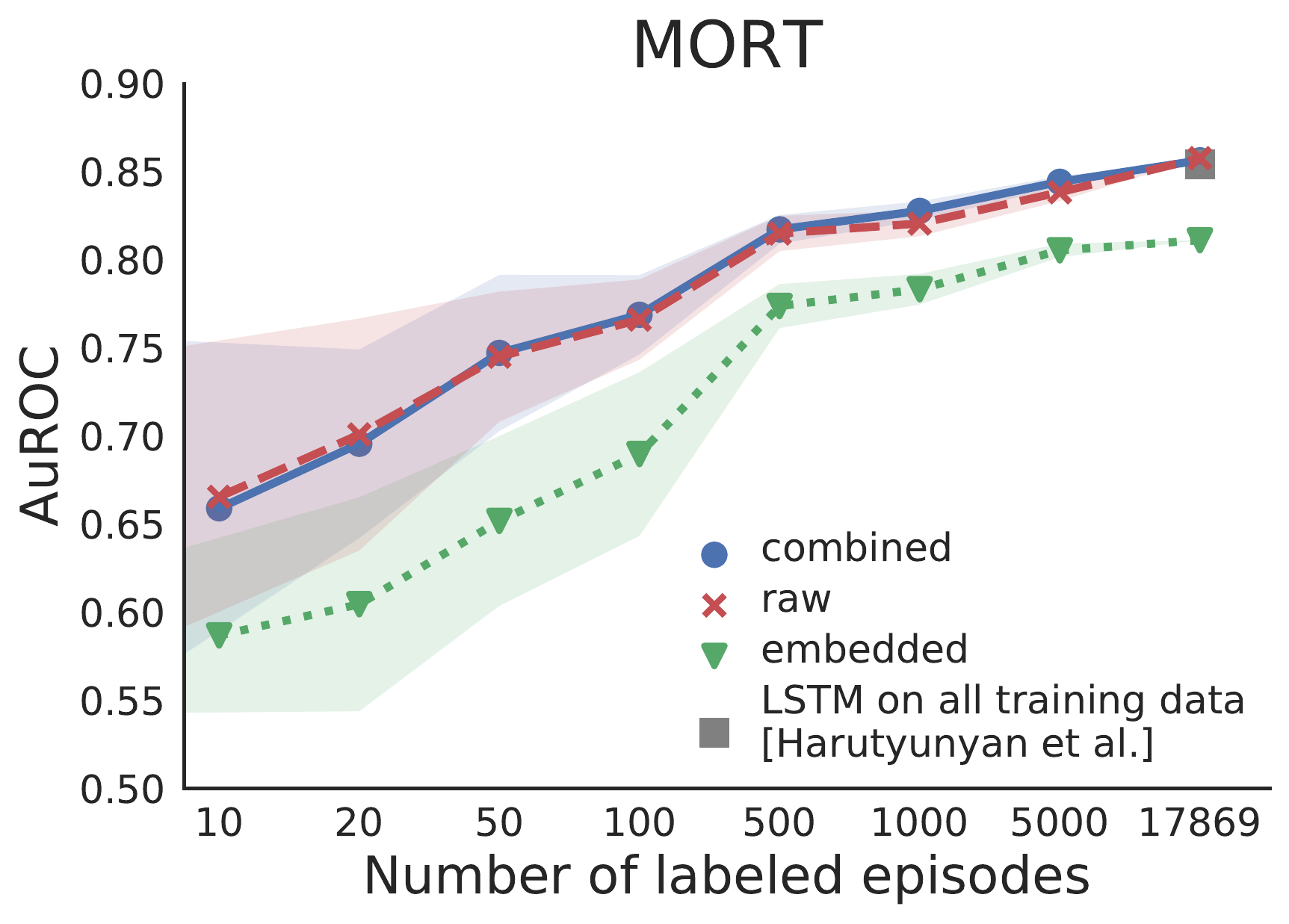}
    \hspace{5mm}
    \includegraphics[width=0.3\textwidth,keepaspectratio]{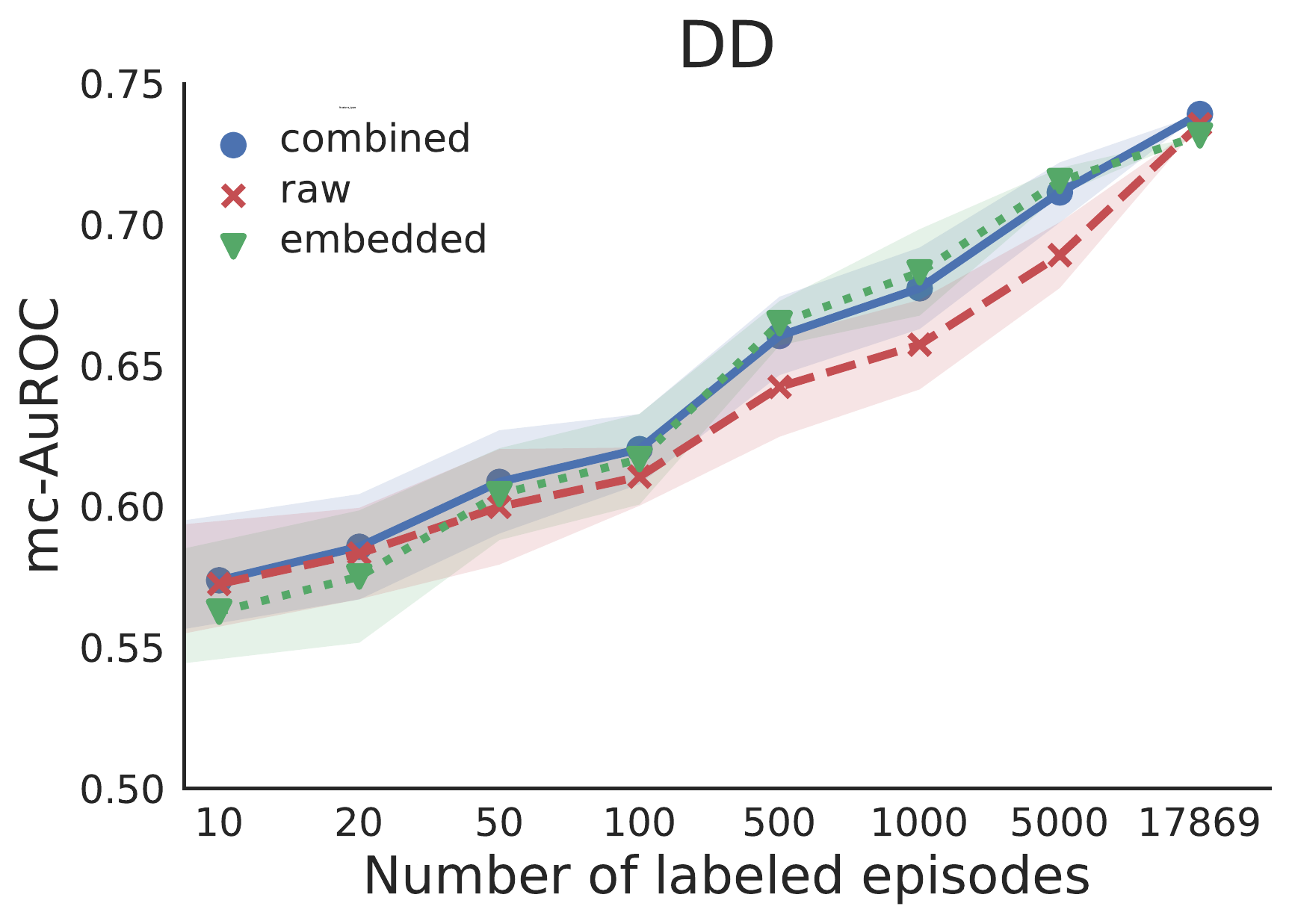}
    \hspace{5mm}
    \includegraphics[width=0.3\textwidth,keepaspectratio]{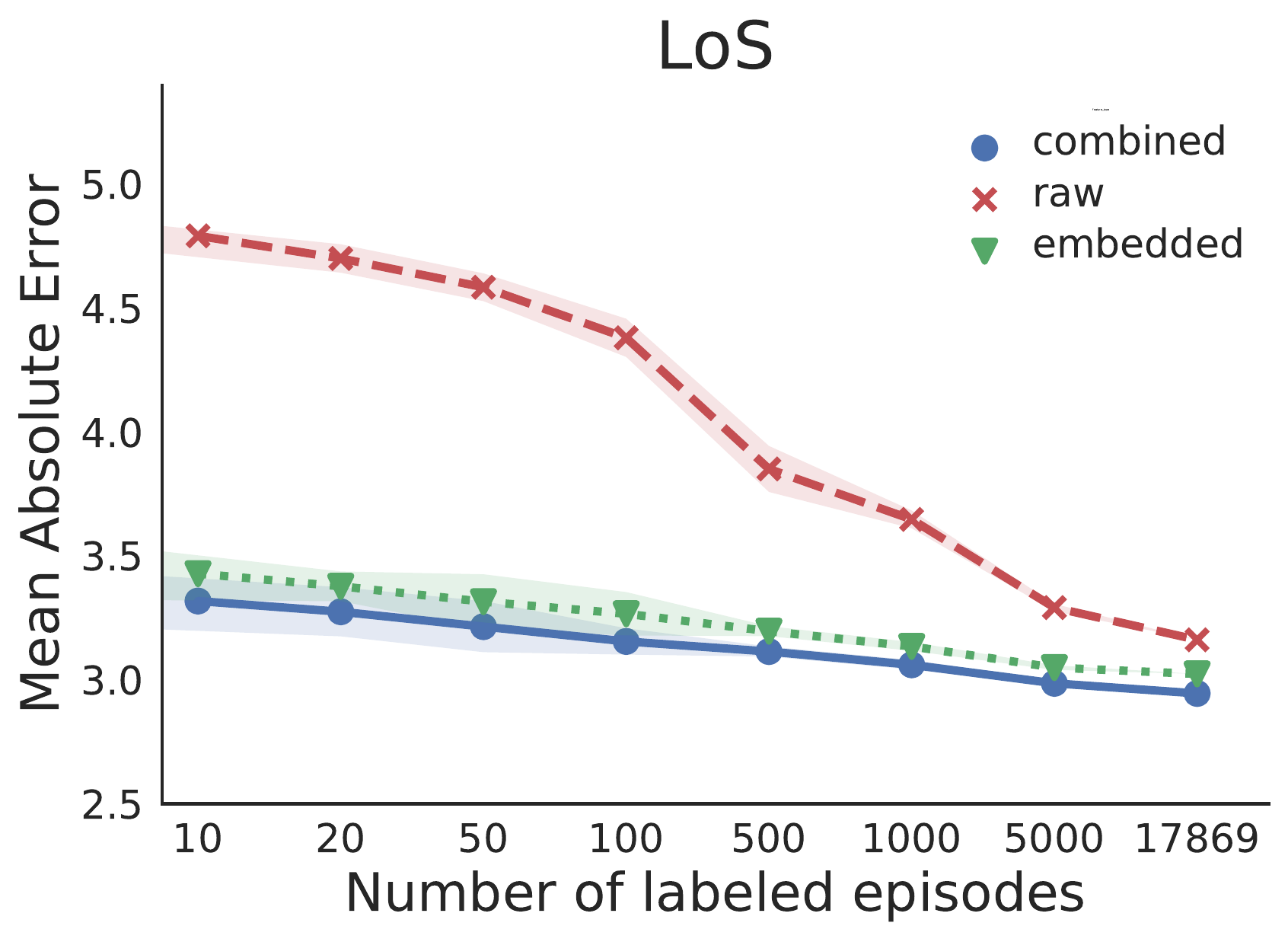}
    
    \vspace{-2mm}
    
    \caption{The respective performance of using only the ``raw'' features,
        only the ``embedded'' episode representations, or the ``combined''
        features for training a logistic (for \mort and \dd) or ridge (for \los) regression model using either the single time
        series modality (top) or all four data modalities (bottom) as input. The x-axis gives the
        number of labeled training episodes (out of a total of $17~869$ episodes
        in the full training set). $20$ random samples of each subset size were
        evaluated for each representation. The markers give the mean
        performance on the entire test set, and the shaded regions show one standard
        deviation above and below the mean. For \mort, we also indicate the
        performance of the best single-task LSTM used in~\cite{Harutyunyan2017} 
        Please note that \emph{higher AuROC values are better} (for \mort and \dd),
        while \emph{lower \mae values are better} (for \los).
        \label{fig:results}}
\vspace{-5mm}
\end{figure*}

We first consider the impact of learning embeddings on only the time
series data present in the original benchmarks for use in downstream 
prediction. Figure~\ref{fig:results}(top)
compares the performance of the \raw baseline (using just the hand-crafted features) compared
to first learning representations of the time series attribute type using \epmd (\embedded). We also include
a third strategy (\combined) in which we combine the original features with the 
representations learned by \epmd as described in Section~\ref{sec:patient-learning}. (Please see \appresulttables for detailed tables of
all results.)

As the figure shows, the embeddings are particularly helpful in small-data
scenarios for \mort and \los. Prior work in the graph-embedding community has observed
similar phenomena, so our findings are in line with that existing work. The
embeddings are particularly well-suited for \los. Indeed, this is the only task 
in this single data modality setting in which \combined meaningfully differs from
\raw.

We observed that \raw exhibits particularly high variance in performance
for small subsample sizes for \los, while the variance for \embedded are in line
with the other settings. This case particularly highlights the ability of \epmd
to reduce noise and improve performance in missing data scenarios, even when only
a single modality is available.

These results also show that both \combined and \raw match state-of-the-art
performance by LSTMs reported in prior work~\cite{Harutyunyan2017}.
Further, the LSTMs were trained in a completely supervised manner;
thus, their performance is expected to deteriorate
when using smaller training sets~\cite{Ramachandran2017}.

\paragraph{Multi-modal data}

We then extend our analysis to the multi-attribute case. As described in
Section~\ref{sec:label-learning}, we use \epmd to learn embeddings for each of the
four attribute types (time series features, text notes, demographics, and episode
identity within the graph) independently. The final representation for the episode
is the concatenation of all attribute type representations.
For \raw, we used standard text (term-frequency inverse document-frequency) and
categorical (one-hot encoding) variable preprocessing.
We construct the graph for \epmd using additional information about the episode
admission (see Section~\ref{sec:preprocessing}). 
We also encode these data as categorical
variables and include them for \raw.

Figure~\ref{fig:results}(bottom) shows the performance after adding the additional
attributes. Relative to their performance in the single-attribute setting,
the performance for all three representations improves across nearly all tasks
and observed episode subsets. For \mort, as in the single-attribute case, \embedded is clearly worse than using
 only the hand-crafted features;
however, combining the embeddings with the raw features does no harm, and 
\combined and \raw perform virtually identically. Both mildly surpass the
state-of-the-art results previously reported~\cite{Harutyunyan2017}.
The \combined representation benefits significantly from the additional data
attributes on the \los task. Indeed, it outperforms both \raw and \embedded
across all subsample sizes. In this task, \combined 
effectively uses information from \emph{both} \raw and \embedded, rather than
simply learning to use the best, as seems the case in other settings.

Finally, all of the representations show significant improvement in the \dd
task when the additional modalities are available. Both \embedded and \combined
significantly outperform \raw for several of the subset sizes.

\paragraph{Statistical significance}

Across all of the sample sizes, \embedded is significantly outperformed by \raw
(paired sample t-test, Benjamini-Hochberg multiple-test corrected \pvalue $< 0.05$)
and \combined on the \dd task in the single-modality settings. Both are also
significantly better than \embedded on \mort in the multiple modality setting. On
the other hand, in the multi-modal setting for \los, \embedded is significantly better than
\raw, and \combined is significantly better than both. \combined is also
significantly better than \raw in the multi-modal case on \dd.

\paragraph{Impact of missing data}

\begin{figure}
    \centering
    \includegraphics[width=0.22\textwidth,keepaspectratio]{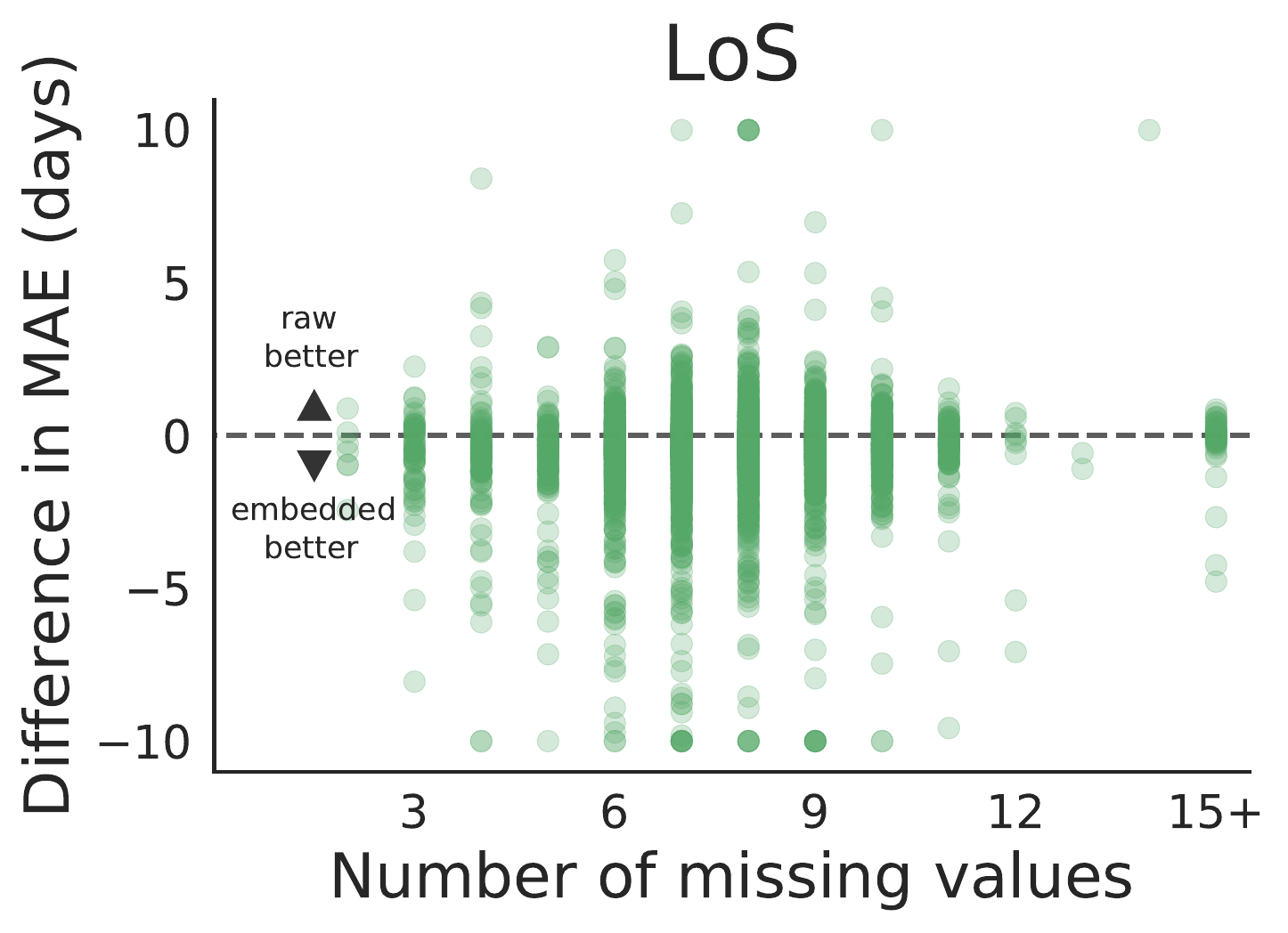}
    \includegraphics[width=0.22\textwidth,keepaspectratio]{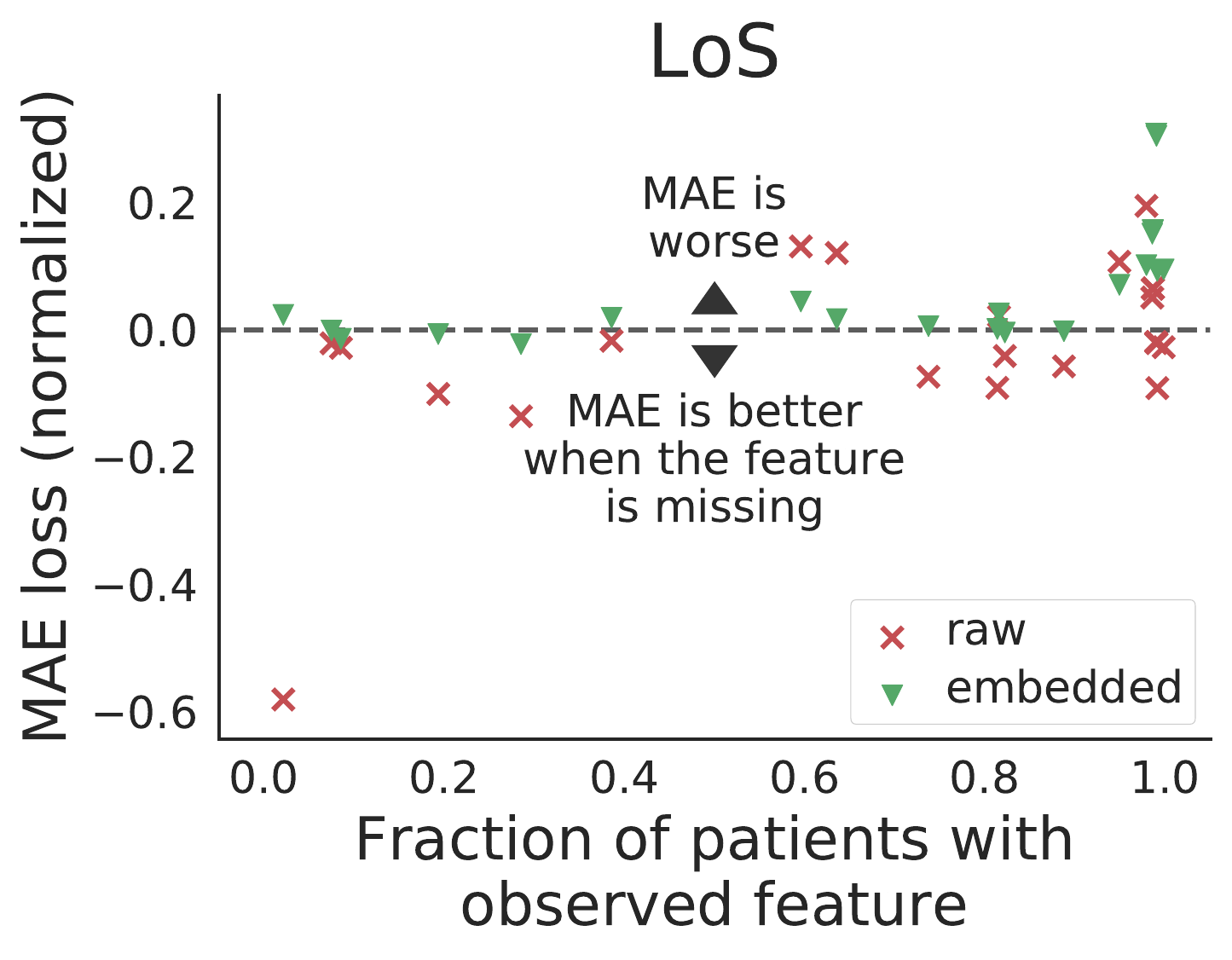}    
    
    \vspace{-2mm}

    \caption{(left) Difference in the mean absolute error on all test instances when using $100$ labeled samples for training. Negative values indicate that \embedded produced better predictions that \raw. (right) Comparison of the MAE when a feature is observed for a test instance compared to when it is missing. The x-axis gives the percentage of patients for which a specific feature was observed; the y-axis shows the difference between the MAE for instances when the feature is observed and when it is not. Each marker corresponds to one feature. A negative value indicates that predictions are more accurate for patients where the feature is missing. \label{fig:missing-data}}
\vspace{-5mm}
\end{figure}

We hypothesize that \embedded outperforms \raw for instances with many missing values.
Figure~\ref{fig:missing-data}(left) confirms that, on the \los task, for patients with few missing values, \embedded and \raw perform similarly.
\embedded outperforms \raw the most in cases with $5$ to $10$ missing values ($\approx 20-40\%$);
\embedded continues to result in more accurate predictions for larger numbers of missing values, but the difference is not as stark.

We next consider the effect on prediction quality when specific features are missing.
As shown in Figure~\ref{fig:missing-data}(right), predictions based on \embedded for \los are similar regardless of whether most features are observed or missing;
this suggests that \epmd effectively propagates useful information about missing features throughout the graph.

On the other hand, \raw features lead to \emph{worse} predictions when most of the less-frequent features are observed.
This helps explain the better performance of \embedded compared to \raw on \los;
\embedded-based predictions effectively use all of the available information, while models trained on \raw are actually misled by rare attributes.

For \raw, missing temperature led to the worst change in performance, while features related to the Glasgow coma scale had the largest negative effect for \embedded. \raw results in much better predictions when the capilary refill rate is missing (the bottom left marker in the plot).
All of the MAE losses are given in \appfeaturemaeloss.

\paragraph{Feature importance}
As previously described, \epmd allows us to identify the original features which most influence predictions.
For \los, we consider the average contribution to the final prediction across all test instances (based on the learned model coefficients and embeddings for each instance) as the importance for each feature.
The plots in \appfeatureimportance show that temperature and pH are the most influential features for \embedded-based predictions; the Glasgow coma scale features are the next most influential.
Interestingly, the contributions of temperature and pH are similar regardless of whether they are observed.
This again suggests that \epmd effectively propagates useful information about missing features.
On the other hand, the Glasgow coma scale features contribute much less when they are not observed.

\section{Related Work}
\label{sec:related-work}

We review some of the most related work on
missing data and representation learning for clinical datasets.

\paragraph{Missing data in clinical datasets}

A large body of recent research has addressed the problem of predicting patient
outcomes from multivariate time series.
The availabiliy of large, high-quality datasets like \mimic~\cite{Johnson2016a}
and those made available through the Computing in Cardiology challenges~\cite{Silva2012}
has surely spurred some of this work. This data entails missingness almost by design since different vital signs are typically observed at different frequencies. 
 Methodologically, much of this work
~\cite{Rajkomar2018,Harutyunyan2017,Lipton2016a,Lipton2016,Che2015,Suresh2016,Hammerla2015}
borrows from and extends recent developments in the deep learning
community, such as  long short-term memory (LSTM) networks~\cite{Hochreiter1997}.
Some Bayesian methods~\cite{Lasko2013,Marlin2012,Lehman2008} have also been developed.
While our approach builds upon this work, such as by extending previously-described hand-crafted
features~\cite{Harutyunyan2017,Lipton2016a}, the use of multi-modal data
is clearly novel in our work compared to this body. Further, our novel message
passing scheme for missing data distinguishes our contribution from this work.

\paragraph{Representation learning and clinical data}

A second approach to predicting patient outcomes from \ehrs has instead focused
on learning representations from discrete observations, such as prescribed
medications and International Classification of Diseases (\icd) codes assigned
during billing. Much of this work~\cite{Choi2016,Choi2016a,Choi2016b} builds on
the recent successes of deep learning to include domain-specific background knowledge. Some
recent work~\cite{Choi2016d} has also considered text modalities, though they 
use  word co-occurrence matrices.

Our use of a graph to explicitly capture similarity relationships among patients
distinguishes our work from these prior approaches. Additionally, these
methods typically do not consider the impact of missing data. In contrast, we explicitly
learn a representation of the missing observations.

\section{Discussion}
\label{sec:discussion}

We have presented an approach for learning representations for ICU episodes from
multimodal data, which could come from \ehrs. Our approach extends an
existing representation learning framework, embedding propagation~\cite{GarciaDuran2017},
to handle missing data. Empirically, we demonstrate that our approach outperforms
a state-of-the-art baseline in predicting hospital visit duration and discharge destination when
multiple attribute types are available; when combined with hand-crafted features,
our approach is competitive with long short-term memory networks for in-hospital
mortality prediction.

Future developments of our approach include extending both the considered tasks
and data included in the model. We did not consider
the ``computational phenotyping'' task~\cite{Lasko2013,Che2015}  in this work.
This is typically formulated as a multi-label classification problem; the
representations learned with \epmd can be combined with standard models for that
setting. Currently, our model does not incorporate medications, procedures and
treatments that occur during an episode; clearly, these have
bearing on the patient outcomes. Since such data is available in \mimic, we plan
to integrate appropriate attribute types for these data.



\bibliographystyle{aaai}
\bibliography{library}

\end{document}


%
\title{Supplement: Learning Representations of Missing Data for\\Predicting Patient Outcomes}
\author{Brandon Malone, Alberto Garc{\'i}a-Dur{\'a}n, and Mathias Niepert}
\maketitle

\section{Appendix A: Time series data}
\label{app:time-series-data}

\begin{itemize}
    \item Capillary refill rate
    \item Diastolic blood pressure
    \item Fraction inspired oxygen
    \item Glascow coma scale eye opening
    \item Glascow coma scale motor response
    \item Glascow coma scale total
    \item Glascow coma scale verbal response
    \item Glucose
    \item Heart Rate
    \item Height
    \item Mean blood pressure
    \item Oxygen saturation
    \item Respiratory rate
    \item Systolic blood pressure
    \item Temperature
    \item Weight
    \item pH
\end{itemize}

\section{Appendix B: Note category assignments}
\label{app:note-categories}

Table~\ref{tab:mapping-mimic} provides the mapping between the \|{CATEGORY} field in the \|{NOTEEVENTS}
table in \mimic and the $6$ types used in this work. 

\begin{table}[H]
\caption{Note type mapping \label{tab:mapping-mimic}}
\centering
\small
\begin{tabular}{l|l}
\|{CATEGORY} & Label                 \\ \hline
Case Management               & NOTE OTHER BOW             \\
Consult                       & NOTE OTHER BOW             \\
Discharge summary             & NOTE DISCHARGE SUMMARY BOW \\
ECG                           & NOTE ECG BOW               \\
Echo                          & NOTE ECHO BOW              \\
General                       & NOTE NURSING BOW           \\
Nursing                       & NOTE NURSING BOW           \\
Nursing/other                 & NOTE NURSING BOW           \\
Nutrition                     & NOTE OTHER BOW             \\
Pharmacy                      & NOTE OTHER BOW             \\
Physician                     & NOTE NURSING BOW           \\
Radiology                     & NOTE RADIOLOGY BOW         \\
Rehab Services                & NOTE OTHER BOW             \\
Respiratory                   & NOTE RESPITORY BOW         \\
Social Work                   & NOTE OTHER BOW            
\end{tabular}
\end{table}

\section{Appendix C: Discharge destinations}
\label{app:discharge-destinations}

Table~\ref{tab:discharge-destinations} gives the mapping between the \|{DISCHARGE{\textunderscore}LOCATION}
field in the \|{ADMISSIONS} table in \mimic and the $6$ types used in this work. 

\begin{table}[H]
\caption{Discharge location mapping \label{tab:discharge-destinations}}
\centering
\small
\begin{tabular}{l|l}
\|{DISCHARGE{\textunderscore}LOCATION}   & \dd label \\ \hline
DEAD/EXPIRED              & MORTALITY{\textunderscore}INHOSPITAL    \\
HOME                      & HOME                     \\
HOME HEALTH CARE          & HOME                     \\
HOME WITH HOME IV PROVIDR & HOME                     \\
HOSPICE-HOME              & HOME                     \\
REHAB/DISTINCT PART HOSP  & REHAB                    \\
SNF                       & SNF                      \\
SNF-MEDICAID ONLY CERTIF  & SNF                      \\
LEFT AGAINST MEDICAL ADVI & LEFT                     \\
LONG TERM CARE HOSPITAL   & TRANSFER                 \\
DISCH-TRAN TO PSYCH HOSP  & TRANSFER                 \\
DISC-TRAN CANCER/CHLDRN H & TRANSFER                 \\
OTHER FACILITY            & TRANSFER                 \\
SHORT TERM HOSPITAL       & TRANSFER                 \\
HOSPICE-MEDICAL FACILITY  & TRANSFER                 \\
ICF                       & TRANSFER                 \\
DISC-TRAN TO FEDERAL HC   & TRANSFER                
\end{tabular}
\end{table}
    
\section{Appendix D: Hyperparameter selection}
\label{app:hyperparameter-selection}

For the results in Section~4 (Experiments) of the main paper, we use an inner three-fold
cross-validation on the training set to select the hyperparameters for all 
linear models. We never use the testing set during training. In particular, the 
search grids we use are given in Table~\ref{tab:search-grid}.

\begin{table}[H]
\caption{Hyperparameter search grid \label{tab:search-grid}}
\centering
Logistic regression (\mort and \dd)
\begin{tabular}{l|l}
Hyperparameter           & Values                        \\ \hline
class weight             & balanced, no change           \\
regularization type      & l1, l2                        \\
regularization parameter & 1.0, 0.1, 0.01, 0.001, 0.0001
\end{tabular}
\end{table}

\begin{table}[H]
\centering
Ridge regression (\los)

\begin{tabular}{l|l}
Hyperparameter           & Values                        \\ \hline
regularization parameter & $10^{-6}$, $10^{-5}$ \dots $10^{+7}$
\end{tabular}
\end{table}

\noindent
Please consult the \sklearn documentation for more details about the hyperparameters.

\section{Appendix E: Result tables}
\label{app:result-tables}

Tables~\ref{tab:first-res} -- \ref{tab:last-res} give the values shown in Figure~3 of the main text.

\begin{table*}
\centering
\caption{\mort, time series attribute type only \label{tab:first-res}}
\begin{tabular}{l|lll}
Number of labeled episodes & \combined                       & \embedded                       & \raw \\ \hline
$10           $ & $0.556 \pm 0.044$ & $0.605 \pm 0.065$ & $0.551 \pm 0.041$ \\
$20           $ & $0.575 \pm 0.047$ & $0.609 \pm 0.057$ & $0.570 \pm 0.066$ \\
$50           $ & $0.679 \pm 0.051$ & $0.673 \pm 0.049$ & $0.673 \pm 0.048$ \\
$100          $ & $0.745 \pm 0.027$ & $0.708 \pm 0.029$ & $0.745 \pm 0.025$ \\
$500          $ & $0.809 \pm 0.010$ & $0.765 \pm 0.010$ & $0.811 \pm 0.011$ \\
$1~000         $ & $0.820 \pm 0.007$ & $0.781 \pm 0.007$ & $0.822 \pm 0.006$ \\
$5~000         $ & $0.840 \pm 0.003$ & $0.797 \pm 0.005$ & $0.840 \pm 0.005$ \\
$17~869        $ & $0.852 \pm 0.000$ & $0.804 \pm 0.000$ & $0.849 \pm 0.000$
\end{tabular}
\end{table*}

\begin{table*}
\centering
\caption{\dd, time series attribute type only}
\begin{tabular}{l|lll}
Number of labeled episodes & \combined                       & \embedded                       & \raw                            \\\hline
$10           $ & $0.539 \pm 0.014$ & $0.537 \pm 0.017$ & $0.545 \pm 0.021$ \\
$20           $ & $0.557 \pm 0.023$ & $0.549 \pm 0.018$ & $0.560 \pm 0.021$ \\
$50           $ & $0.588 \pm 0.016$ & $0.566 \pm 0.016$ & $0.590 \pm 0.015$ \\
$100          $ & $0.607 \pm 0.013$ & $0.584 \pm 0.015$ & $0.608 \pm 0.014$ \\
$500          $ & $0.645 \pm 0.012$ & $0.618 \pm 0.011$ & $0.644 \pm 0.013$ \\
$1~000         $ & $0.658 \pm 0.014$ & $0.634 \pm 0.011$ & $0.656 \pm 0.016$ \\
$5~000         $ & $0.688 \pm 0.009$ & $0.659 \pm 0.007$ & $0.688 \pm 0.009$ \\
$17~869$        & $	0.709 \pm 0.000$ & $0.673 \pm 0.000$ & $0.707 \pm 0.000$
\end{tabular}
\end{table*}

\begin{table*}
\centering
\caption{\los, time series attribute type only}
\begin{tabular}{l|lll}
Number of labeled episodes & \combined                       & \embedded                       & \raw             \\\hline
$10           $ & $4.672 \pm 1.657$ & $3.416 \pm 0.129$ & $5.242 \pm 2.612$ \\
$20           $ & $4.714 \pm 1.396$ & $3.416 \pm 0.100$ & $5.143 \pm 1.503$ \\
$50           $ & $3.972 \pm 1.443$ & $3.263 \pm 0.073$ & $3.978 \pm 1.443$ \\
$100          $ & $3.508 \pm 0.190$ & $3.230 \pm 0.091$ & $3.509 \pm 0.190$ \\
$500          $ & $3.334 \pm 0.313$ & $3.166 \pm 0.064$ & $3.399 \pm 0.339$ \\
$1~000         $ & $3.202 \pm 0.088$ & $3.131 \pm 0.017$ & $3.239 \pm 0.089$ \\
$5~000         $ & $3.001 \pm 0.038$ & $3.053 \pm 0.006$ & $3.034 \pm 0.036$ \\
$17~869        $ & $2.956 \pm 0.000$ & $3.028 \pm 0.000$ & $2.985 \pm 0.000$
\end{tabular}
\end{table*}

\begin{table*}
\centering
\caption{\mort, All attribute types}
\begin{tabular}{l|lll}
Number of labeled episodes  & \combined                       & \embedded                       & \raw \\\hline
$10           $ & $0.659 \pm 0.096$ & $0.587 \pm 0.044$ & $0.665 \pm 0.082$ \\
$20           $ & $0.696 \pm 0.053$ & $0.605 \pm 0.061$ & $0.701 \pm 0.066$ \\
$50           $ & $0.747 \pm 0.044$ & $0.652 \pm 0.048$ & $0.745 \pm 0.037$ \\
$100          $ & $0.769 \pm 0.023$ & $0.690 \pm 0.046$ & $0.766 \pm 0.023$ \\
$500          $ & $0.817 \pm 0.008$ & $0.774 \pm 0.012$ & $0.815 \pm 0.010$ \\
$1~000         $ & $0.828 \pm 0.005$ & $0.783 \pm 0.009$ & $0.821 \pm 0.007$ \\
$5~000         $ & $0.844 \pm 0.003$ & $0.805 \pm 0.004$ & $0.838 \pm 0.005$ \\
$17~869        $ & $0.856 \pm 0.000$ & $0.811 \pm 0.000$ & $0.858 \pm 0.000$
\end{tabular}
\end{table*}

\begin{table*}
\centering
\caption{\dd, All attribute types}
\begin{tabular}{l|lll}
Number of labeled episodes & \combined                       & \embedded                       & \raw \\\hline
$10           $ & $0.574 \pm 0.019$ & $0.563 \pm 0.020$ & $0.573 \pm 0.020$ \\
$20           $ & $0.586 \pm 0.019$ & $0.575 \pm 0.023$ & $0.583 \pm 0.016$ \\
$50           $ & $0.609 \pm 0.018$ & $0.604 \pm 0.016$ & $0.600 \pm 0.020$ \\
$100          $ & $0.620 \pm 0.012$ & $0.617 \pm 0.016$ & $0.611 \pm 0.010$ \\
$500          $ & $0.660 \pm 0.014$ & $0.665 \pm 0.008$ & $0.642 \pm 0.018$ \\
$1~000         $ & $0.677 \pm 0.014$ & $0.683 \pm 0.015$ & $0.657 \pm 0.016$ \\
$5~000         $ & $0.711 \pm 0.011$ & $0.715 \pm 0.005$ & $0.689 \pm 0.012$ \\
$1~7869        $ & $0.739 \pm 0.000$ & $0.732 \pm 0.000$ & $0.736 \pm 0.000$
\end{tabular}
\end{table*}

\begin{table*}
\centering
\caption{\los, All attribute types \label{tab:last-res}}
\begin{tabular}{l|lll}
Number of labeled episodes & \combined                       & \embedded                       & \raw                            \\\hline
$10           $ & $3.318 \pm 0.109$ & $3.429 \pm 0.106$ & $4.792 \pm 0.055$ \\
$20           $ & $3.275 \pm 0.099$ & $3.378 \pm 0.060$ & $4.701 \pm 0.057$ \\
$50           $ & $3.215 \pm 0.103$ & $3.316 \pm 0.111$ & $4.584 \pm 0.056$ \\
$100          $ & $3.155 \pm 0.053$ & $3.267 \pm 0.088$ & $4.380 \pm 0.077$ \\
$500          $ & $3.115 \pm 0.020$ & $3.197 \pm 0.020$ & $3.851 \pm 0.093$ \\
$1~000         $ & $3.061 \pm 0.019$ & $3.135 \pm 0.018$ & $3.648 \pm 0.036$ \\
$5~000         $ & $2.986 \pm 0.011$ & $3.050 \pm 0.009$ & $3.292 \pm 0.015$ \\
$17~869        $ & $2.945 \pm 0.000$ & $3.024 \pm 0.000$ & $3.162 \pm 0.000$
\end{tabular}
\end{table*}

\section{Appendix F: Feature MAE loss}
\label{app:feature-mae-loss}

Table~\ref{tab:mae-loss} gives the normalized MAE loss shown in Figure~4(right) of the main text.

\begin{table*}[]
\caption{MAE loss shown in Figure~4(right) of the main text. \label{tab:mae-loss}}
\begin{tabular}{l|ll}
Attribute                              & Embedded MAE loss & Raw MAE loss \\ \hline
Capillary refill rate                & -0.023301         & 0.580816     \\
Diastolic blood pressure             & -0.156282         & -0.064640    \\
ETHNICITY                              & 0.002419          & 0.057705     \\
Fraction inspired oxygen             & 0.022698          & 0.136004     \\
Glasgow coma scale eye opening     & -0.307864         & 0.019429     \\
Glasgow coma scale motor response  & -0.304350         & 0.020988     \\
Glasgow coma scale total            & -0.044169         & -0.130732    \\
Glasgow coma scale verbal response & -0.307864         & 0.019429     \\
Glucose                                & -0.094687         & 0.027558     \\
Heart Rate                            & -0.156282         & -0.064640    \\
Height                                 & 0.006909          & 0.100878     \\
MARITAL STATUS                        & -0.070432         & -0.106718    \\
Mean blood pressure                  & -0.156282         & -0.064640    \\
NOTE DISCHARGE SUMMARY BOW          & 0.233990          & 0.284654     \\
NOTE ECG BOW                         & -0.025505         & -0.021364    \\
NOTE ECHO BOW                        & -0.018772         & 0.017316     \\
NOTE NURSING BOW                     & -0.016769         & -0.120556    \\
NOTE OTHER BOW                       & 0.001341          & 0.021422     \\
NOTE RADIOLOGY BOW                   & 0.004466          & 0.041344     \\
NOTE RESPITORY BOW                   & 0.014245          & 0.028457     \\
Oxygen saturation                     & -0.092130         & 0.091678     \\
Respiratory rate                      & -0.150736         & -0.050517    \\
Systolic blood pressure              & -0.156282         & -0.064640    \\
Temperature                            & -0.101287         & -0.194498    \\
Weight                                 & -0.005509         & 0.073845     \\
pH                                     & -0.001081         & 0.091238    
\end{tabular}
\end{table*}

\section{Appendix G: Feature importance}
\label{app:feature-importance}

\begin{figure}
    \centering
    \includegraphics[width=0.4\textwidth,keepaspectratio]{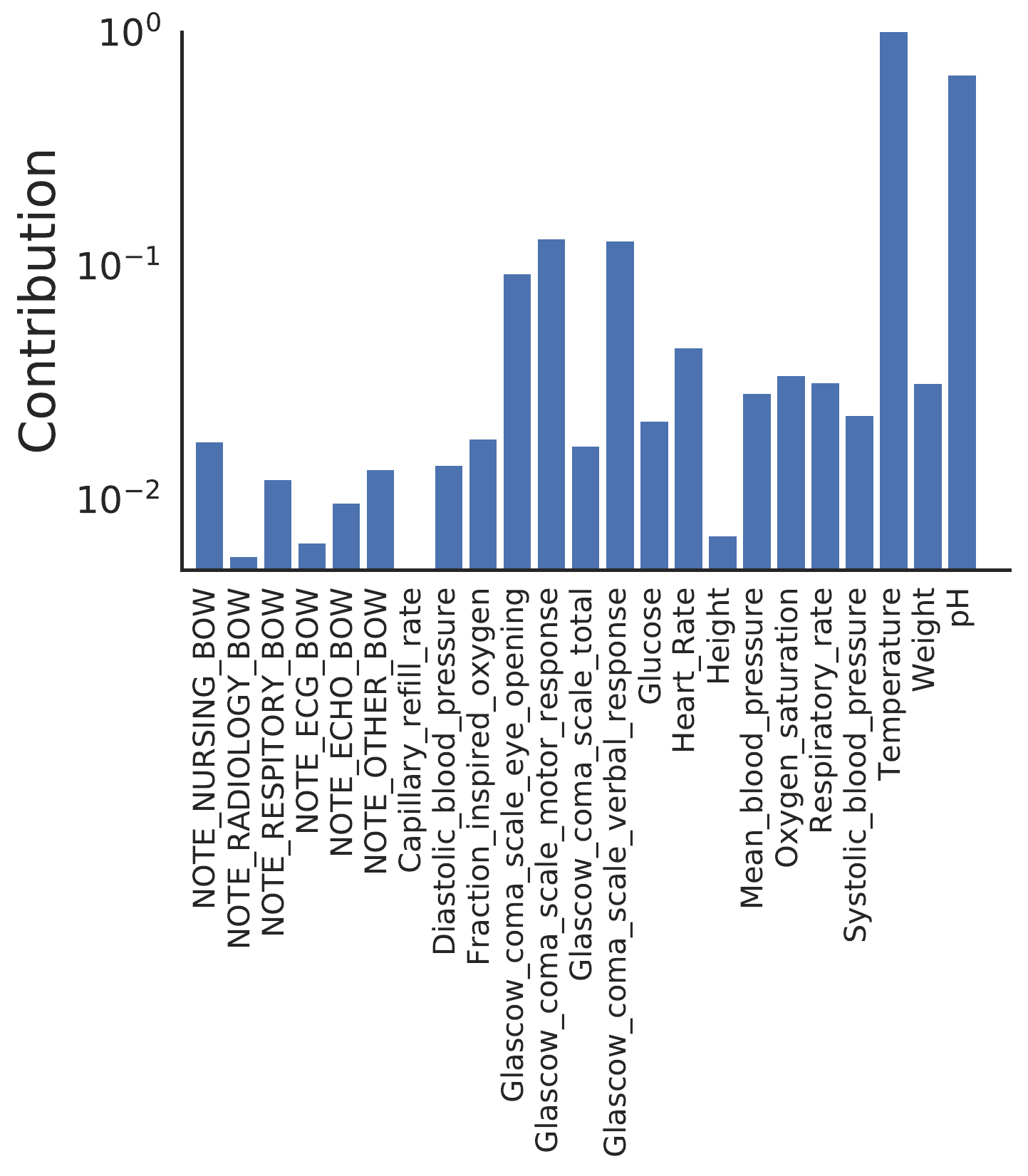}
    \caption{The average contribution of each feature to final predictions across all test instances on \los when trainined with $100$ labeled instances. \label{fig:feature-contribution}}
\end{figure}

Figure~\ref{fig:feature-contribution} shows the average contribution of each feature to the final prediction for \los when training on $100$ labeled instances.
Specifically, the contribution for each feature on a particular instance is calculated as the sum of the element-wise product of the embedding for the feature with the appropriate coefficients from the learned linear model;
we additionally take the absolute value to consider only the magnitude of the contribution.

\begin{figure}
    \centering
    \includegraphics[width=0.4\textwidth,keepaspectratio]{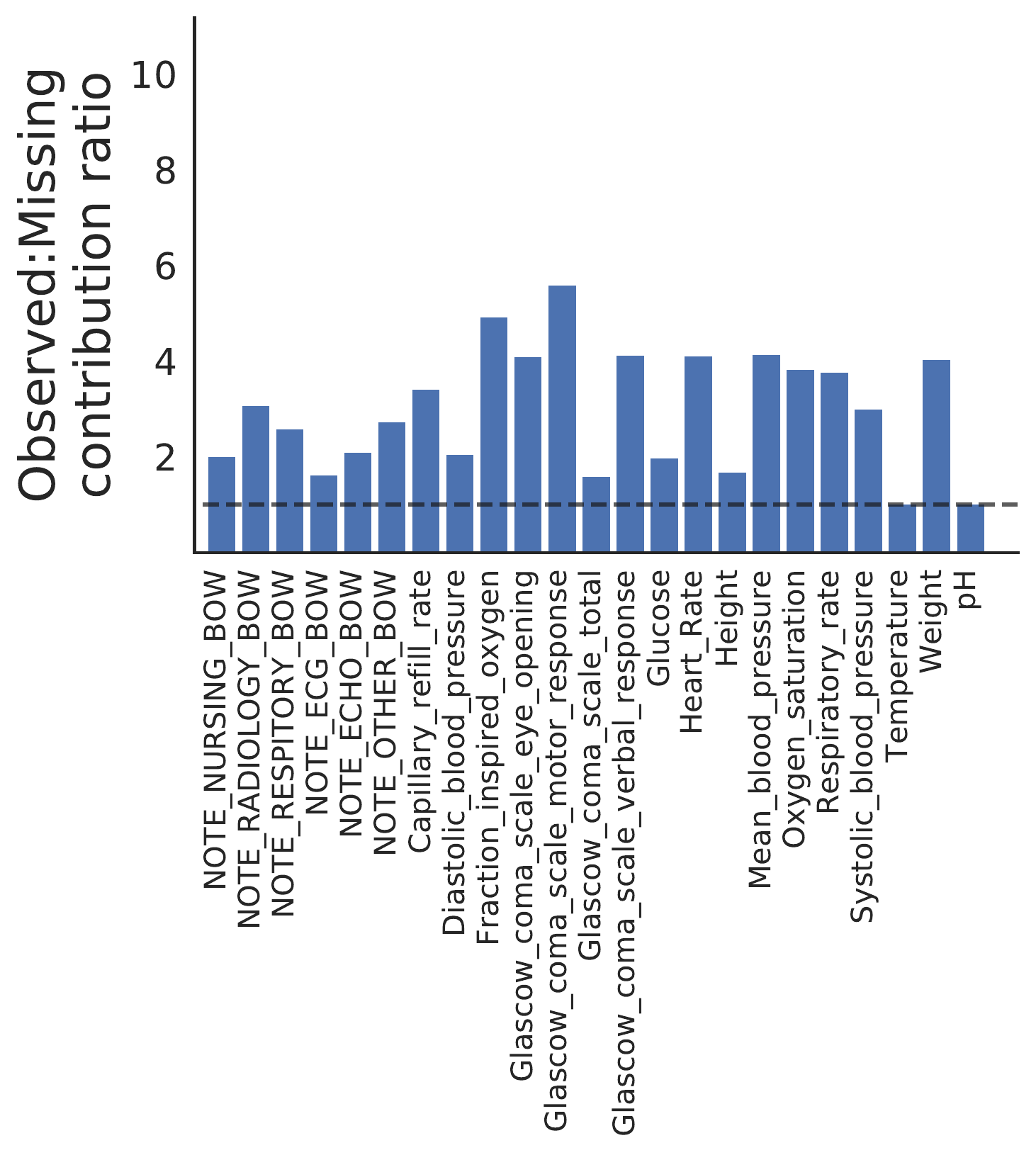}
    \caption{The ratio of the average contribution of each feature to final predictions across all test instances when the feature is observed compared to when it is missing. The dashed line indicates where the ratio is $1$, which means the feature contributes equally whether it is observed or not. Higher values indicate that the feature contributes more when it is observed compared to when it is missing. \label{fig:feature-contribution-ratio}}
\end{figure}

Figure~\ref{fig:feature-contribution-ratio} shows the ratio of the average contributions for each feature when it is observed compared to when it is missing.
Larger values indicate that the feature contributes more when it is observed than when it is missing.
For example, the embedded Glasgow coma scale motor response feature contributes, on average, about $6$ times more to the final prediction when it is observed than when it is missing.
In contrast, temperature and pH contribute roughly the same regardless of whether they are observed or not.

